\documentclass[a4paper,twoside]{article}

\usepackage{epsfig}
\usepackage{subcaption}
\usepackage{calc}
\usepackage{amssymb}
\usepackage{amstext}
\usepackage{amsmath, amsfonts}
\usepackage{amsthm}
\usepackage{multicol}
\usepackage{pslatex}
\usepackage{apalike}
\usepackage{xcolor}         
\usepackage{multirow}
\usepackage{url}            
\usepackage{booktabs}       
\usepackage[bottom]{footmisc}
\usepackage{SCITEPRESS}     
\usepackage{pifont}
\newcommand{\cmark}{\ding{51}}%
\newcommand{\xmark}{\ding{55}}%
\usepackage{hyperref}

\begin{document}

\title{Dynamically Modular and Sparse General Continual Learning}

\author{\authorname{Arnav Varma\sup{1}, Elahe Arani\sup{\dagger1,2} and Bahram Zonooz\sup{\dagger1,2}}
\affiliation{\sup{1}Advanced Research Lab, NavInfo Europe, Eindhoven, The Netherlands}
\affiliation{\sup{2}Department of Mathematics and Computer Science, Eindhoven University of Technology, The Netherlands}
\email{arnav.varma@navinfo.eu, e.arani@tue.nl, bahram.zonooz@gmail.com}
{\scriptsize \sup{\dagger} Contributed equally}
}


\abstract{
    Real-world applications often require learning continuously from a stream of data under ever-changing conditions. When trying to learn from such non-stationary data, deep neural networks (DNNs) undergo catastrophic forgetting of previously learned information. Among the common approaches to avoid catastrophic forgetting, rehearsal-based methods have proven effective. However, they are still prone to forgetting due to task-interference as all parameters respond to all tasks. To counter this, we take inspiration from sparse coding in the brain and introduce dynamic modularity and sparsity (\textit{Dynamos}) for rehearsal-based general continual learning. In this setup, the DNN learns to respond to stimuli by activating relevant subsets of neurons. We demonstrate the effectiveness of \textit{Dynamos} on multiple datasets under challenging continual learning evaluation protocols. Finally, we show that our method learns representations that are modular and specialized, while maintaining reusability by activating subsets of neurons with overlaps corresponding to the similarity of stimuli. The code is available at \url{https://github.com/NeurAI-Lab/DynamicContinualLearning}. 
    }

\onecolumn \maketitle \normalsize \setcounter{footnote}{0} \vfill

\section{\uppercase{Introduction}}
\label{sec:intro}

Deep neural networks (DNNs) have achieved human-level performance in several applications~\cite{greenwald2021whole,taigman2014deepface}. 
These networks are trained on the multiple tasks within an application with the data being received under an independent and identically distributed (i.i.d.) assumption. This assumption is satisfied by shuffling the data from all tasks and balancing and normalizing the samples from each task in the application~\cite{hadsell2020embracing}. Consequently, DNNs can achieve human-level performance on all tasks in these applications by modeling the joint distribution of the data as a stationary process. Humans, on the other hand, can model the world from inherently non-stationary and sequential observations~\cite{french1999catastrophic}. Learning continually from the more realistic sequential and non-stationary data is crucial for many applications such as lifelong learning robots~\cite{thrun1995lifelong} and self-driving cars~\cite{nose2019study}. However, vanilla gradient-based training for such continual learning setups with a continuous stream of tasks and data leads to task interference in the DNN's parameters, and consequently, catastrophic forgetting on old tasks~\cite{mccloskey1989catastrophic,kirkpatrick2017overcoming}.
Therefore, there is a need for methods to alleviate catastrophic forgetting in continual learning.

Previous works have aimed to address these challenges in continual learning. These can be broadly classified into three categories.
First, regularization-based methods~\cite{kirkpatrick2017overcoming,schwarz2018progress,zenke2017continual}  that penalize changes to the parameters of DNNs to reduce task interference.
Second, parameter isolation methods~\cite{DBLP:conf/iclr/Adel0T20} that assign distinct subsets of parameters to different tasks.
Finally, rehearsal-based methods~\cite{DBLP:conf/iclr/ChaudhryRRE19} that co-train on current and stored previous samples. 
Among these, regularization-based and parameter isolation-based methods often require additional information (such as task-identity at test time and task-boundaries during training), or unconstrained growth of networks. These requirements fail to meet general continual learning (GCL) desiderata~\cite{delange2021continual,farquhar_towards_2018}, making these methods unsuitable for GCL.

Although rehearsal-based methods improve over other categories and meet GCL desiderata, 
they still suffer from catastrophic forgetting through task interference in the DNN parameters, as all parameters respond to all examples and tasks.
This could be resolved by inculcating task or example specific parameter isolation in the rehearsal-based methods. However, it is worth noting that unlike parameter isolation methods, modularity and sparsity in the brain is not static. There is evidence that the brain responds to stimuli in a dynamic and sparse manner, with different modules or subsets of neurons responding "dynamically" to different stimuli~\cite{graham2006sparse}.
The advantages of a dynamic and sparse response to stimuli have been explored in deep learning in stationary settings through mechanisms such as gating of modules~\cite{veit2018convolutional}, early-exiting~\cite{li2017not,tkhu2019triplewins}, and dynamic routing~\cite{wang2018skipnet}, along with training losses that incentivize sparsity of neural activations~\cite{wu2018blockdrop}. These studies observed that DNNs trained to predict dynamically also learn to respond differently to different inputs. Furthermore, the learned DNNs demonstrate clustering of parameters in terms of tasks such as similarity, difficulty, and resolution of inputs~\cite{wang2018skipnet,veit2018convolutional}, indicating dynamic modularity.
Hence, we hypothesize that combining rehearsal-based methods with dynamic sparsity and modularity could help further mitigate catastrophic forgetting in a more biologically plausible fashion while adhering to GCL desiderata.

To this end, we propose Dynamic Modularity and Sparsity (\textit{Dynamos}), a general continual learning algorithm that combines rehearsal-based methods with dynamic modularity and sparsity. Concretely, we seek to achieve three objectives: dynamic and sparse response to inputs with specialized modules, competent performance, and reducing catastrophic forgetting. To achieve dynamic and sparse responses to inputs, we define multiple agents in our DNN, each responsible for dynamically zeroing out filter activations of a convolutional layer based on the input to that layer. The agents are rewarded for choosing actions that remove activations (sparse responses) if the network predictions are accurate, but are penalized heavily for choosing actions that lead to inaccurate predictions. Agents also rely on prototype losses to learn specialized features. To reduce forgetting and achieve competent performance, we maintain a constant-size memory buffer in which we store previously seen examples. The network is retrained on previous examples alongside current examples to both maintain performance on current and previous tasks, as well as to enforce consistency between current and previous responses to stimuli. \textit{Dynamos} demonstrates competent performance on multiple continual learning datasets under multiple evaluation protocols, including general continual learning. 
Additionally, our method demonstrates similar and overlapping responses for similar inputs and disparate responses for dissimilar inputs. Finally, we demonstrate that our method can simulate the trial-to-trial variability observed in humans~\cite{faisal2008noise,werner1963variability}.

\section{\uppercase{Related Work}}
\label{sec:related_work}
Research in deep learning has approached the dynamic compositionality and sparsity observed in the human brain through dynamic neural networks, where different subsets of neurons or different sub-networks are activated for different stimuli~\cite{bengio2015conditional,bolukbasi2017adaptive}.
This can be achieved through early exiting~\cite{tkhu2019triplewins}, dynamic routing through mixtures of experts or multiple branches~\cite{collier2020routing,wang2022learning}, and through gating of modules~\cite{wang2018skipnet}. 
Early-exiting might force the DNN to learn specific features in its earlier layers and consequently hurt performance~\cite{wu2018blockdrop} as the earlier layers of DNNs are known to learn general purpose features~\cite{yosinski2014transferable}. 
Dynamic routing, on the other hand, would require the growth of new experts in response to new tasks that risk unconstrained growth, or the initialization of a larger DNN with branches corresponding to the expected number of tasks~\cite{chen2020mitigating}. Dynamic networks with gating mechanisms, meanwhile, have been shown to achieve competent performance in i.i.d. training with standard DNNs embedded with small gating networks~\cite{veit2018convolutional,wu2018blockdrop,wang2018skipnet}. These gating networks emit a discrete keep/drop decision for each module, depending on the input to the module or the DNN. As this operation is non-differentiable, a Gumbel Softmax approximation~\cite{veit2018convolutional,wang2018skipnet}, or an agent trained with policy gradients~\cite{wu2018blockdrop,sutton2018reinforcement} is commonly used in each module to enable backpropagation.
However, unlike the latter, the Gumbel-Softmax approximation induces an asymmetry between the forward pass activations at inference and training~\cite{wang2018skipnet}. Furthermore, these methods are not applicable to continual learning.

Recent works have attempted to build dynamic networks for continual learning setups~\cite{chen2020mitigating,abati2020conditional}, where data arrive in a more realistic sequential manner. 
InstAParam~\cite{chen2020mitigating}, Random Path Selection (RPS)~\cite{rajasegaran2019random}, and MoE~\cite{collier2020routing} start with multiple parallel blocks at each layer, finding input-specific or task-specific paths within this large network. Nevertheless, this requires knowledge of the number of tasks to be learned ahead of training. More importantly, initializing a large network might be unnecessary as indicated by the competent performance of dynamic networks with gating mechanisms in i.i.d training.
In contrast to this, MNTDP~\cite{veniat2021efficient}, LMC~\cite{ostapenko2021continual}, and CCGN~\cite{abati2020conditional} start with a standard architecture and grow units to respond to new data or tasks. Of these, MNTDP and LMC develop task-specific networks where all inputs from the same task elicit the same response and therefore do not show a truly dynamic response to stimuli. CCGN, however, composes convolutional filters dynamically to respond to stimuli, using a task-specific vector for every convolutional filter, and task boundaries to freeze frequently active filters. However, this leads to unrestrained growth and fails in the absence of task-boundaries, 
which makes it unsuitable for general continual learning.

Therefore, we propose a general continual learning method with dynamic modularity and sparsity (Dynamos) induced through reinforcement learning agents trained with policy gradients.

\begin{figure*}[t]
\centering
  \includegraphics[width=\linewidth]{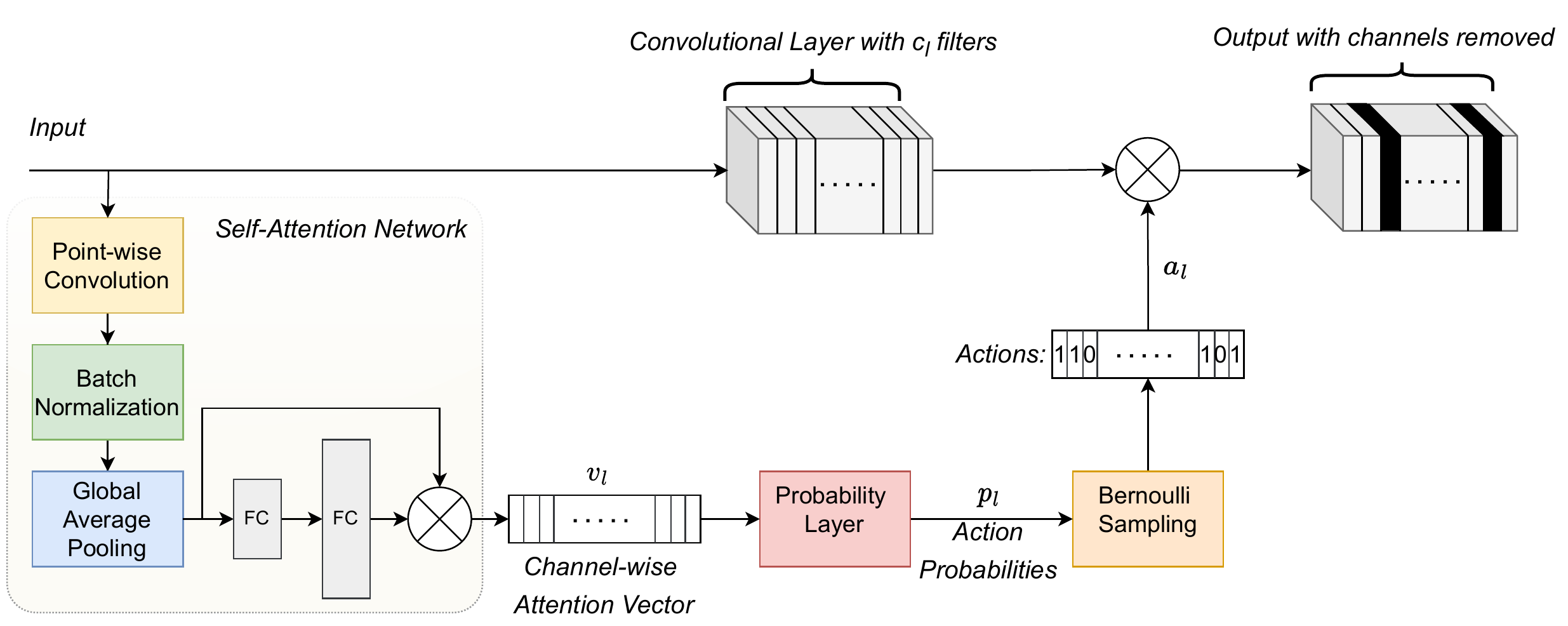}
  \caption{An overview of \textit{Dynamos}' dynamic and sparse response mechanism at the $l^{\text{th}}$ convolutional layer. Blacked activations are removed. The agent (bottom path) self-attention network uses a pointwise convolution to match output channels and global average pooling to get a channel-length flattened vector. This is sent through an MLP with one hidden layer and Sigmoid activation, and multiplied with the original channel-length representation to get the channel-wise self-attention vector.} 
\label{fig:architecture}
\end{figure*}

\section{\uppercase{Methodology}}
\label{sec:methodology}
Humans learn continually from inherently non-stationary and sequential observations of the world without catastrophic forgetting, even without supervision about tasks to be performed or the arrival of new tasks, maintaining a bounded memory throughout.
This involves, among other things, making multi-scale associations between current and previous observations~\cite{goyal2020inductive} and responding sparsely and dynamically to stimuli~\cite{graham2006sparse}. The former concerns consolidation of previous experiences and ensuring that learned experiences evoke a similar response. The latter concern dynamically composing a subset of the specialized neural modules available to respond to stimuli, reusing only the relevant previously learned information. This also avoids erasure of information irrelevant to current stimuli but relevant to previous experiences.
We now formulate an approach for dynamic sparse and modular general continual learning that mimics these procedures with DNNs.

\subsection{Dynamic, Modular, and Sparse response to stimuli}
\label{subsec:agents}
To achieve a dynamic, modular, and sparse response to inputs, we use a DNN $F$ with a policy to compose a subset of the available modules in each layer to respond to the input to that layer. More specifically, we use a CNN which is incentivized to drop some channels in its activations adaptively using policy gradients~\cite{sutton2018reinforcement,williams1992simple}.

Let us consider the $l^{\text{th}}$ convolutional layer with $c_{l}$ output channels $\forall l \in \{1, 2,...L\}$, where $L$ is the total number of convolutional layers in the network. The input to the convolutional layer is processed using an agent module with actions $a_l \in \{0, 1\}^{c_l}$ as output, where each action represents the decision to drop (action = $0$) or keep (action = $1$) the corresponding channel of the output of the convolutional layer. 
The agent module uses a self-attention network to obtain a channel-wise attention vector $v_l$ of dimension $c_{l}$, which is converted into "action probabilities" using a probability layer. The policy for choosing actions is then sampled from a $c_l$-dimensional Bernoulli distribution; 
\begin{equation}
\label{eq:policy}
\begin{split}
    p_l &= \sigma(v_l) \\
    \pi_l(a_l) &= \prod_{i=1}^{c_{l}} p_{l, i}^{a_{l, i}}(1 - p_{l, i})^{(1 - a_{l, i})},
\end{split}
\end{equation}
where $p_l \in (0, 1)^{c_{l}}$ is the output of the probability layer $\sigma$, and $\pi_l$ is the policy function. The final output of the convolutional layer is the channel-wise product of the actions with the output of the convolution. This policy formulation is used at each convolutional layer in the CNN, leading to $L$ agents in total. The overall structure of an agent for a convolutional layer is shown in Figure~\ref{fig:architecture}.

These agents are rewarded for dropping channels while making accurate predictions through a reward function. For an input to the DNN $X$ applied to classification with label $Y$:
\begin{equation}
\label{eq:fwd}
    \begin{split}
        Z, V &= F(X) \text{, } V= [v_1 | v_2, ... | v_L] \\
        \hat{Y} &= \arg \max Z,
    \end{split}
\end{equation}
where $Z$ refers to the logits.
Now, the ratio of activations or channels that were retained in the layer $l$ is determined by $\frac{1}{c_l}\sum_{i=1}^{c_l}a_{l, i}$. So, for a target activation retention rate per layer or "keep ratio" $kr$, the reward function is as follows:
\begin{equation}
\label{eq:reward}
R_l(X, Y) = 
\begin{cases}
-(kr - \frac{1}{c_l}\sum_{i=1}^{c_l}a_{l, i})^2, &\text{if } \hat{Y} = Y\\
-\lambda(kr - \frac{1}{c_l}\sum_{i=1}^{c_l}a_{l, i})^2, &\text{otherwise.}
\end{cases}
\end{equation}
Therefore, when the DNN's predictions are correct, each agent is rewarded for dropping enough activations to match the "keep ratio" from its corresponding convolutional layer. However, when the prediction is incorrect, each agent is penalized for the same, scaled by a constant penalty factor $\lambda$. The global nature of the reward function, achieved through dependence on the correctness of the prediction, also enforces coordination between agents. Following REINFORCE~\cite{williams1992simple}, the loss from all agents $t=1,2,...L$ is:
\begin{equation}
\label{eq:rwd_loss}
\begin{split}
    L_{R}(X, Y) &= \mathbb{E}_l\mathbb{E}_\pi[-R_l(X, Y)\log\pi_l(a_l)] \\
    &= \mathbb{E}_l\mathbb{E}_\pi[-R_l(X, Y)\log \prod_{i=1}^{c_l}p_{l, i}a_{l, i} \\ &\quad\quad\quad\quad + (1 - p_{l, i})(1 - a_{l, i})] \\
    &= \mathbb{E}_l\mathbb{E}_\pi[-R_l(X, Y)\sum_{i=1}^{c_l}\log[p_{l, i}a_{l, i} \\ &\quad\quad\quad\quad + (1 - p_{l, i})(1 - a_{l, i})]]. \\
\end{split}
\end{equation}

Although the agents along with this loss ensure sparse and dynamic responses from the DNN, they do not explicitly impose any specialization of compositional neural modules seen in humans. 
As the channel-wise "modules" activated in the DNN are directly dependent on the channel-wise attention vectors, we finally apply a specialization loss that we call prototype loss to them. Concretely, for classification, in any batch of inputs, we pull the vectors belonging to the same class together while pushing those from different classes away. This would cause different subsets of channel-wise modules to be used for inputs of different classes. When combined with a sufficiently high "keep ratio", this will encourage overlap and therefore, reuse of relevant previously learned information (for example, reusing channels corresponding to a learned class for a newly observed class) and, consequently, learning of general-purpose features by the modules. For an input batch $X$ with corresponding labels $Y$, and the corresponding batch of concatenated channel-wise attention vectors $V$  (Equation~\ref{eq:fwd}), the prototype loss is given by: 
\begin{equation}
\label{eq:protoype_loss}
    L_{P}(X, Y) = \frac{1 + \Sigma_{(V_1, V_2) \in V^2: Y_1 = Y_2}MSE(V_1, V_2)}{1 + \Sigma_{(V_1, V_2) \in V^2, Y_1 \neq Y_2}MSE(V_1, V_2)},
\end{equation}
where $MSE$ refers to the Mean Squared Error estimator. Note that we only apply this loss to samples for which the predictions were correct. 

\subsection{Multi-Scale associations}
\label{subsec:losses}
As discussed earlier, one of the mechanisms employed by humans to mitigate forgetting is multi-scale associations between current and previous experiences. 

With this goal in mind, we follow recent rehearsal-based approaches~\cite{buzzega2020dark,DBLP:conf/iclr/RiemerCALRTT19} that comply with GCL and use a memory buffer during training to store previously seen examples and responses. The buffer is updated using reservoir sampling~\cite{vitter1985random}, which helps to approximate the distribution of the samples seen so far~\cite{isele2018selective}. However, we only consider the subset of batch samples on which the prediction was made correctly for addition to the memory buffer. These buffer samples are replayed through the DNN alongside new samples with losses that associate the current response with the stored previous response, resulting in consistent responses over time. 

Let $M$ denote the memory buffer and $D_T$ denote the current task stream, from which we sample batches $(X_M, Y_M, Z_M, V_M)$ and $(X_t, Y_t)$, respectively. Here, $Z_M$ and $V_M$ are the saved logits and channel-wise attention vectors corresponding to $X_M$ when it was initially observed. The consistency losses associated with current and previous responses are obtained during the task $T$ as follows:
\begin{equation}
\label{eq:consistency_losses}
\begin{split}
    Z^{'}_M, V^{'}_M &= F(X_M) \\
    L_{C}(Z_M, Z^{'}_M) &= 
    \mathbb{E}_{X_M}[\lVert Z_M - Z^{'}_M \rVert^2_2] \\
    L_{C}(V_M, V^{'}_M) &= 
    \mathbb{E}_{X_M}[\lVert V_M - V^{'}_M \rVert^2_2]. \\
\end{split}
\end{equation}

In addition to consistency losses, we also enforce accuracy, and dynamic sparsity and modularity on the memory samples. Therefore, we have four sets of losses:
\begin{itemize}
    \item Task performance loss on current and memory samples to ensure correctness on current and previous tasks. For classification, we use cross-entropy loss ($L_{CE}$).
    \item Reward losses (Equation~\ref{eq:rwd_loss}) on current and memory samples to ensure dynamic modularity and sparsity on current and previous tasks.
    \item Prototype losses (Equation~\ref{eq:protoype_loss}) on current and memory samples to ensure the specialization of modules on current and previous tasks.
    \item Consistency losses (Equation~\ref{eq:consistency_losses}) for multi-scale associations between current and previous samples.
\end{itemize}
Putting everything together, the total loss becomes:
\begin{equation}
\begin{split}
\label{eq:total_loss}
    L_{\text{total}} &= L_{CE}(X_B, Y_B) + 
    \gamma L_{r}(X_B) 
    \\ &+\beta [L_{CE}(X_M, Y_M) + \gamma L_{r}(X_M)] 
    \\ &+ \alpha L_{C}(Z_M, Z^{'}_M) + \alpha_p L_{C}(V_M, V^{'}_M)
    \\ &+w_p [L_P(X_B, Y_B) + L_P(X_M, Y_M)]. 
\end{split}
\end{equation}

The weights given to the losses - $\alpha$, $\alpha_p$, $\beta$, 
$w_p$, and $\gamma$, and the penalty for misclassification ($\lambda$) and keep ratio ($kr$) in Equation~\ref{eq:reward}, are hyperparameters. Note that we employ a warm-up stage at the beginning of training, where neither the memory buffer nor the agents are employed. This is equivalent to training using only the cross-entropy loss for this period, while the agents are kept frozen. This gives agents a better search space when they start searching for a solution.
We call our method as described above Dynamic modularity and sparsity - \textit{Dynamos}.

\section{\uppercase{Experiment Details}}
\label{subsec:settings}

\paragraph{Datasets.} 
We show results on sequential variants of MNIST~\cite{lecun1998gradient} and SVHN: Seq-MNIST and Seq-SVHN~\cite{netzer2011reading}, respectively. Seq-MNIST and Seq-SVHN divide their respective datasets into $5$ tasks, with $2$ classes per task. Furthermore, to test the applicability of \textit{Dynamos} under general continual learning, we also use the MNIST-360 dataset~\cite{buzzega2020dark}.

\paragraph{Architecture.} 
We use a network based on the ResNet-18~\cite{he2016deep} structure by removing the later two of its four blocks and reducing the number of filters per convolutional layer from $64$ to $32$. The initial convolution is reduced to $3\times3$ to work with smaller image sizes. For the baseline experiments, we did not use any agents. For our method, while agents can be used for all convolutional layers, we only use agents in the second block. We make this choice based on recent studies that observe that earlier layers undergo minimal forgetting~\cite{davari2022probing}, are highly transferrable~\cite{yosinski2014transferable}, and are used for most examples even when learned with dynamic modularity~\cite{abati2020conditional}. We use a sigmoid with a temperature layer as the probability layer in the agents and a probability of $0.5$ as a threshold for picking actions, i.e., channels during inference. The temperature serves the purpose of tuning the range of outputs of the self-attention layers, ensuring that the probabilities being sampled to choose the actions are not too small and that enough activations are chosen to enable learning. The exact network structure used for each experiment, including the self-attention networks of the agents, can be found in Appendix, in Table~\ref{tab:agent} and Table~\ref{tab:arch}. 

\paragraph{Settings.}All methods are implemented in the Mammoth repository\footnote{\url{https://github.com/aimagelab/mammoth/}} in PyTorch $1.6$ and were trained on Nvidia V100 GPUs. The hyperparameters corresponding to each experiment can be found in Appendix, Table~\ref{tab:settings}. We always maintain a keep ratio higher than $1 / Num\_tasks$ to allow the learning of overlapping, reusable, and general-purpose modules. The temperature of the Sigmoid activation of the probability layers is kept at $0.15$ unless mentioned otherwise.

\section{\uppercase{Results}}
\label{sec:results}
We will evaluate \textit{Dynamos} under two standard evaluation protocols that adhere to the core desiderata of GCL.

\subsection{Class-Incremental Learning (CIL)}
\label{subsec:benchmark}
\begin{figure}[t]
\centering
  \includegraphics[width=.9\linewidth]{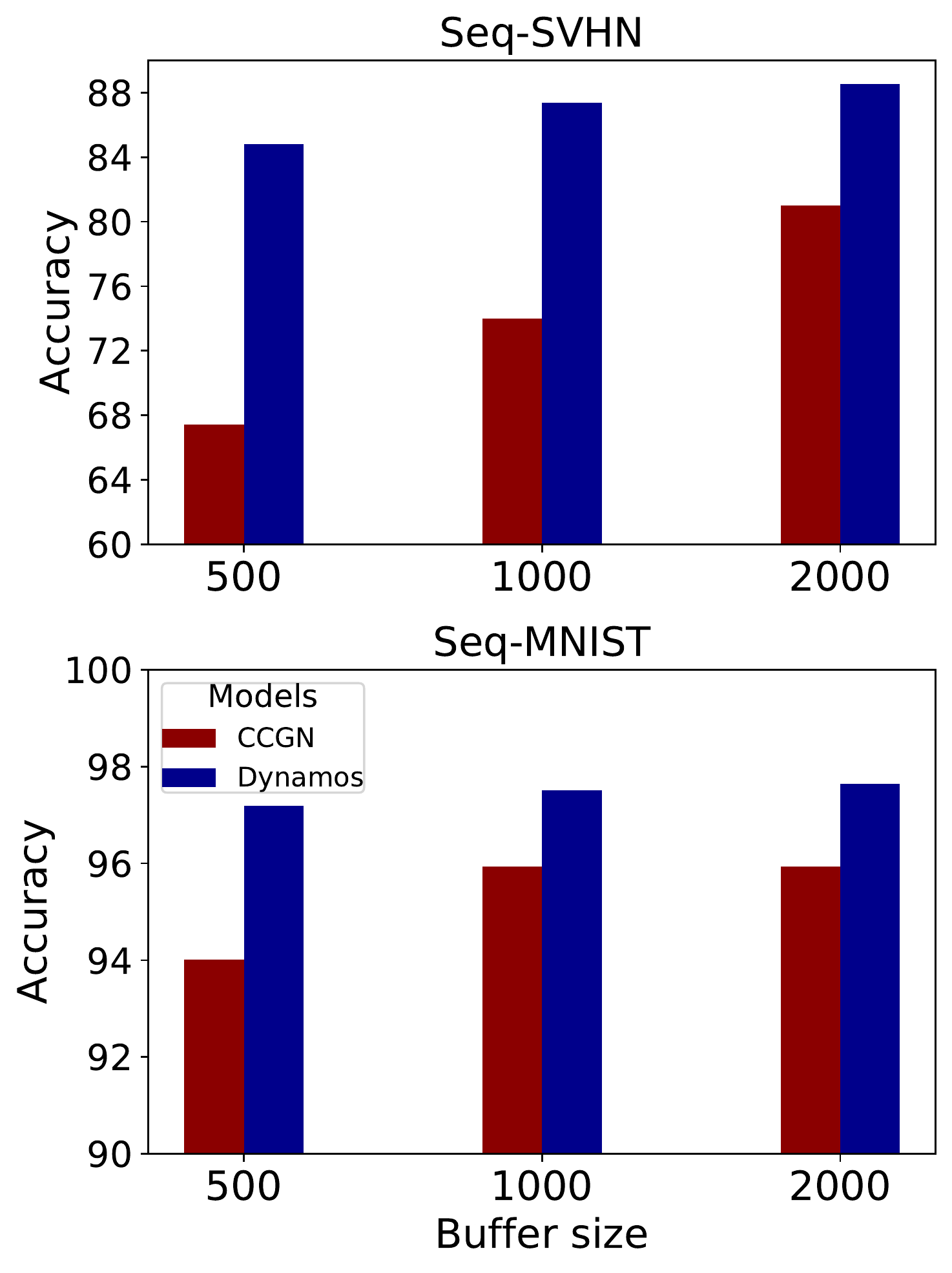}
  \caption{Quantitative results under Class-Incremental Learning protocol. Results are averaged across three seeds. CCGN values taken from the original paper. The precise accuracies can be found in Table~\ref{tab:cil}. } 
\label{fig:cil}
\end{figure}

Class-incremental learning (CIL) refers to the evaluation protocol in which mutually exclusive sets of classes are presented sequentially to the network, and the identity of the task is not provided at the test time, which meets the core desiderata of GCL~\cite{farquhar_towards_2018}. 
We compare against Conditional Convolutional Gated Network (CCGN)~\cite{abati2020conditional}, which also dynamically composes convolutional filters for continual learning.
We observe in Figure~\ref{fig:cil} that \textit{Dynamos} shows higher accuracies on both the Seq-MNIST and Seq-SVHN datasets under all buffer sizes. 
However, CCGN requires a separate task vector for every task per convolutional layer, resulting in unrestricted growth during training, whereas we maintain a bounded memory through training. 
Furthermore, unlike CCGN, we do not leverage the task boundaries or the validation set during training.
Therefore, \textit{Dynamos} outperforms the previous state-of-the-art for dynamic compositional continual learning in class-incremental learning, while showing bounded memory consumption during training.

\subsection{General Continual Learning (GCL)}
\label{subsec:gcl}
\begin{table*}[tb]
\centering
\caption{General continual learning results for multiple buffer sizes. All results are averaged across five seeds.}
\label{tab:gcl}
\begin{tabular}{cc|ccc}
\toprule
\multirow{2}{*}{\begin{tabular}[c]{@{}l@{}}Multi-Scale\\ Associations\end{tabular}} & \multirow{2}{*}{\begin{tabular}[c]{@{}l@{}}Dynamic \\ Modularity\end{tabular}} & \multicolumn{3}{c}{Buffer Size} \\ \cmidrule{3-5} 
&  & 100  & 200 & 500 \\ \midrule
\cmark & \cmark & $\textbf{64.418} \pm 4.095$ & $\textbf{79.638} \pm 2.853$ & $\textbf{90.519} \pm 0.737$ \\
\cmark & \xmark  & $61.192 \pm 3.072$ & $75.364 \pm {1.259}$ & $88.150 \pm 0.888$ \\ 
\hline
\xmark & \xmark & \multicolumn{3}{c}{$18.712\pm0.690$} \\ \bottomrule
\end{tabular}
\end{table*}

\begin{figure*}[t]
\centering
\begin{subfigure}[b]{0.49\textwidth}
    \includegraphics[width=\textwidth]{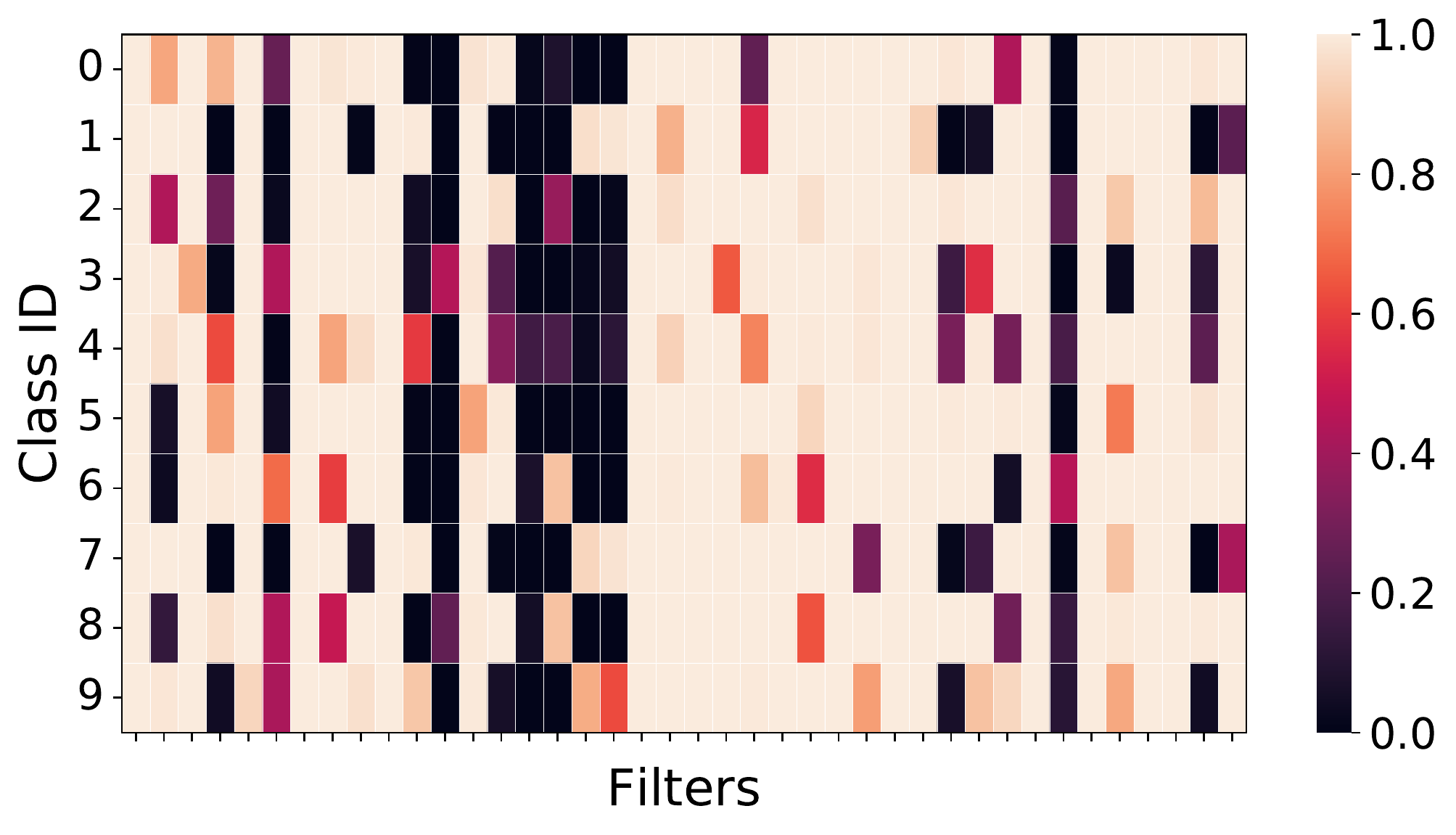}
    \caption{Classwise}
    \label{fig:classwise}
\end{subfigure}
\begin{subfigure}[b]{0.49\textwidth}
    \includegraphics[width=\textwidth]{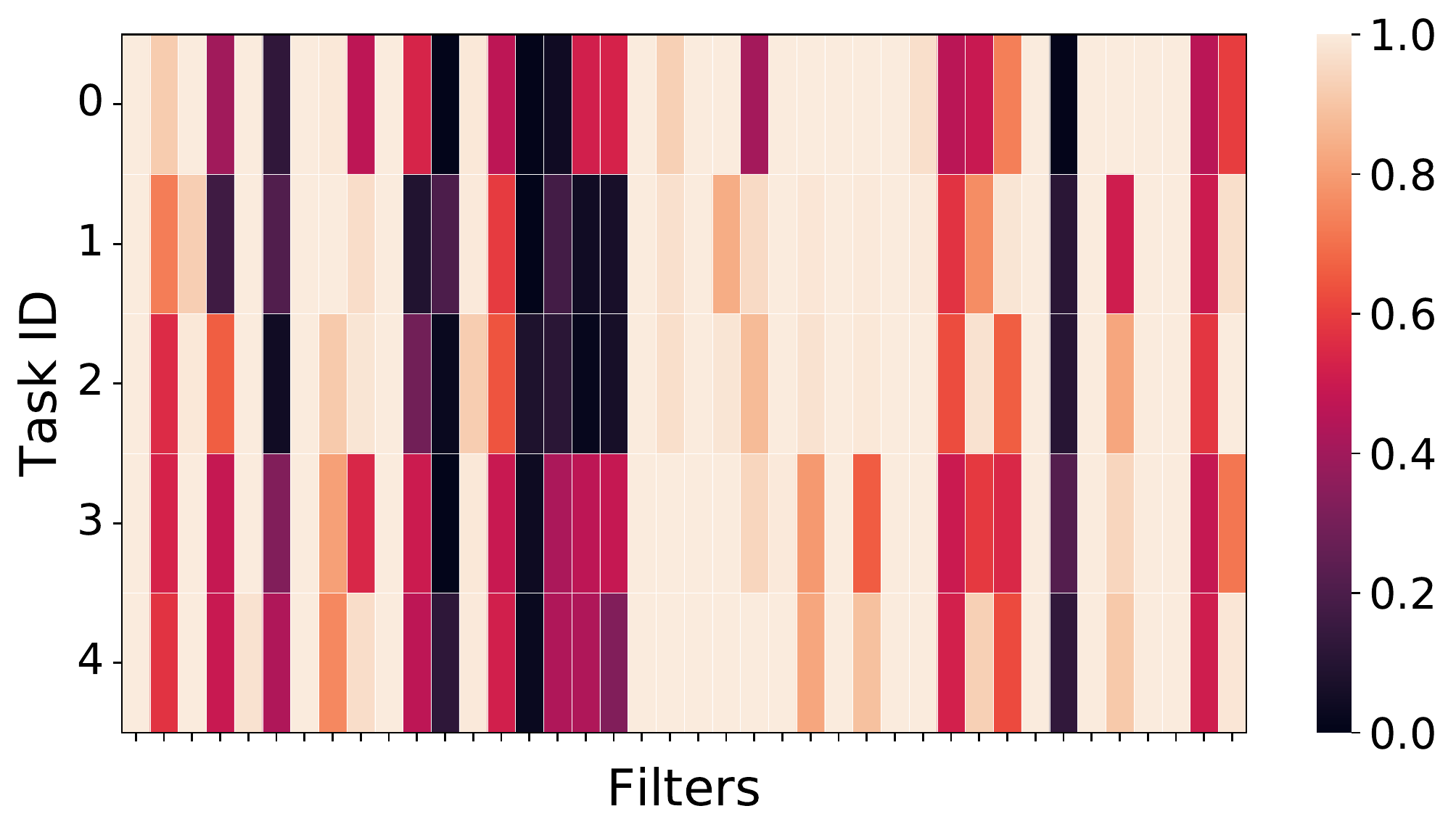}
    \caption{Taskwise}
    \label{fig:taskwise}
\end{subfigure}
\caption{Filter activation rates on the test set for each filter with respect to tasks and classes. For ease of visualization, we only look at the last 40 filters. Full visualizations can be found in Appendix (Figure~\ref{fig:modularity_full}).} 
\label{fig:modularity}
\end{figure*}

\begin{figure}[t]
    \centering
    \includegraphics[width=\columnwidth]{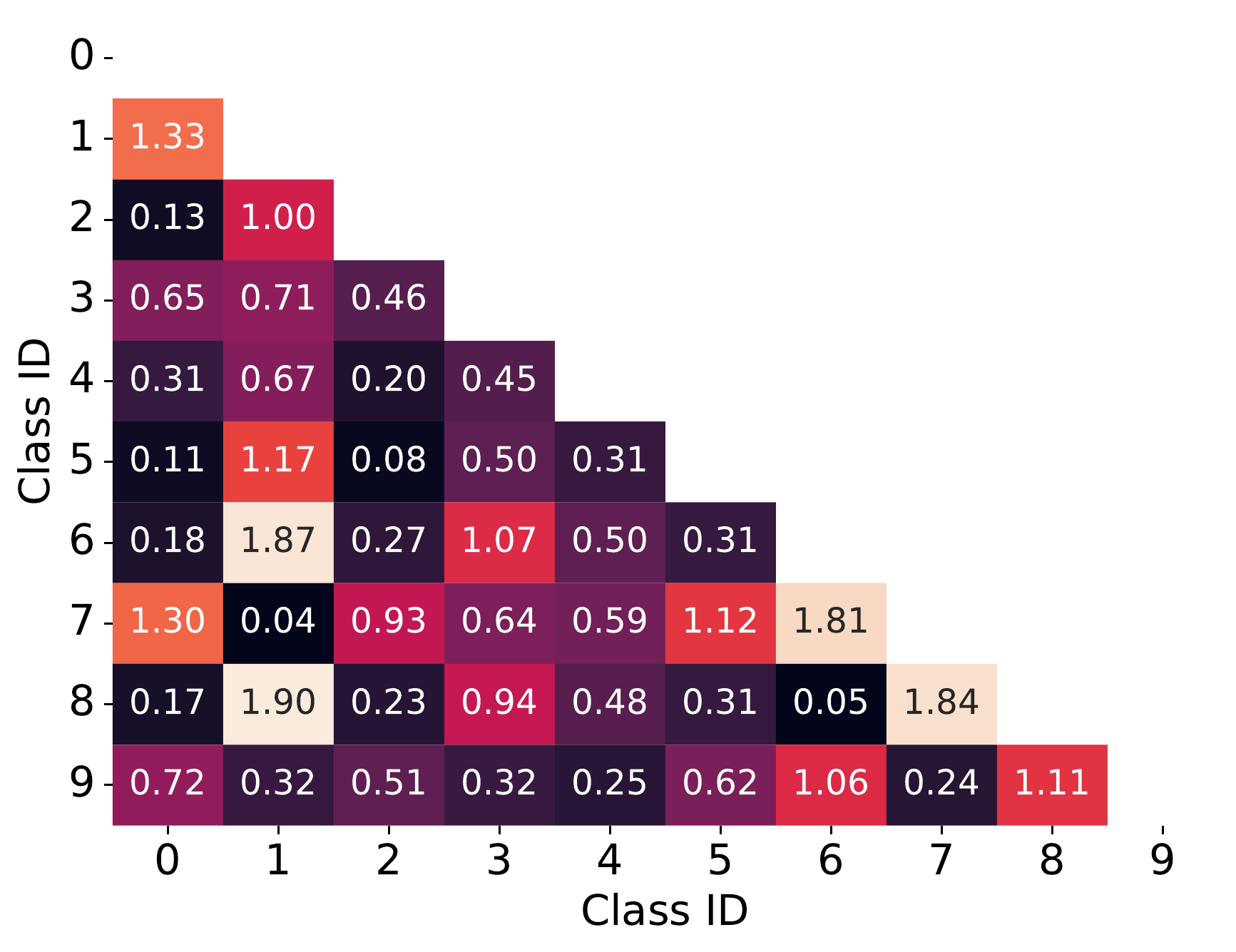}
    \caption{
    Jensen-Shanon Divergences ($\times 100$) of the activation rates of class pairs on the test set.}
    \label{fig:jsd}
\end{figure}

So far, we have observed \textit{Dynamos} under the CIL protocol. Unlike CIL, real-world data streams without clear task boundaries, where the same data may reappear under different distributions (e.g. different poses). Following~\cite{buzzega2020dark}, we approximate this setting using MNIST-360, where tasks overlap in digits (i.e. classes), reappear under different rotations (i.e. distributions), and each example is seen exactly once during training. This serves as a verification of the adherence to the GCL desiderata~\cite{farquhar_towards_2018,delange2021continual}.
We study the impact of both dynamic modularity as well as multi-scale associations by removing them incrementally from \textit{Dynamos}. When neither is used, the learning is done using vanilla gradient-based training, with no strategy to counter forgetting. When dynamic modularity is removed, the learning strategy forms our baseline, where no agents are used, simplifying the total training loss from Equation~\ref{eq:total_loss} to:
\begin{equation}
\label{eq:base_loss}
\begin{aligned}
        L_{\text{base}} = L_{CE}(X_B, Y_B) + \beta L_{CE}(X_M, Y_M) + \\
        \alpha L_{C}(Z_M, Z^{'}_M). 
\end{aligned}
\end{equation}
Table~\ref{tab:gcl} shows that \textit{Dynamos} outperforms the baseline in all buffer sizes, proving that dynamic modularity is advantageous in GCL. Furthermore, when multi-scale associations are also removed, no buffer is used, and the DNN undergoes catastrophic forgetting.
Thus, \textit{Dynamos} is applicable to general continual learning, with dynamic modularity improving over the baseline. We hypothesize that dynamic modularity makes dealing with the blurred task boundaries of GCL easier by adaptively reusing relevant previously learned information, which in this case corresponds to learned filters.

\section{\uppercase{Model Characteristics}}
\label{subsec:analysis}
We now analyze some of the characteristics and advantages of \textit{Dynamos}. For all experiments in this section, we use our model trained on Sequential-MNIST with buffer size $500$. 

\subsection{Dynamic Modularity and Compositionality}
\label{subsubsec:modularity}

Humans show modular and specialized responses to stimuli\cite{meunier2010modular} with dynamic and sparse response to inputs~\cite{graham2006sparse} - a capability that we instilled in our DNN while learning a sequence of tasks by dynamically removing channel activations of convolutional layers. Therefore, we examine the task- and class-wise tendencies of the firing rates of each neuron (filter) in Figure~\ref{fig:modularity}.

It can be seen that \textit{Dynamos} learns a soft separation of both tasks and classes, as evidenced by the per-task and per-class firing rates, respectively, of each filter. This is in contrast to static methods, where all filters react to all examples. Figure~\ref{fig:classwise} further shows that this allows learning of similar activation patterns for similar examples. For example, MNIST digit pairs $1$ and $7$, and $6$ and $8$, which share some shape similarities, also share similarities in their activation patterns/rates. This could be attributed to being able to reuse and drop learned filters dynamically, which causes the DNN to react similarly to similar inputs, partitioning its responses based on example similarities. Additionally, the ability to dynamically reuse filters allows DNNs to learn overlapping activation patterns for dissimilar examples and classes, instead of using completely disparate activation patterns. This also facilitates the learning of sequences of tasks without having to grow the DNN capacity or having a larger capacity at initialization, as opposed to the static parameter isolation methods for continual learning. 

\begin{figure}[t]
    \centering
    \includegraphics[scale=0.4]{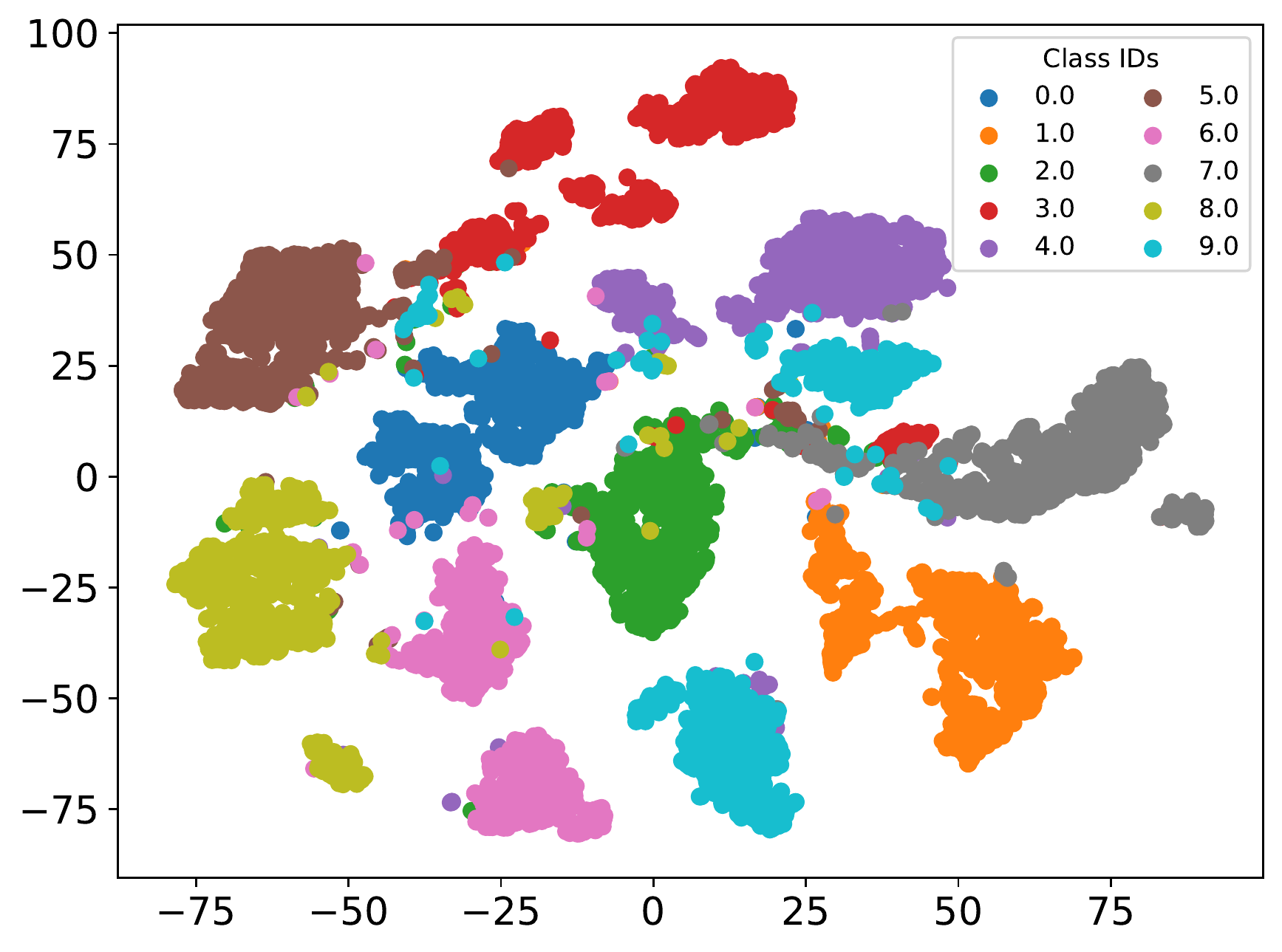}
    \caption{
    t-SNEs on the test set of class prototypes learned from channel-wise self-attention vectors for all classes.}
    \label{fig:clusters}
\end{figure}

Following~\cite{abbasi2022sparsity}, we quantify the overlap between the activation rates for each class pair in the final layer using the Jensen-Shanon divergence (JSD) between them in Figure~\ref{fig:jsd}. Lower JSDs signify higher overlap. The JSD is lowest for the class pair $(1, 7)$ (both digits look like vertical lines), and is $\sim \frac{1}{15^{\text{th}}}$ the average JSD across class pairs, and $\sim \frac{1}{42^{\text{th}}}$ that of the least overlapping class pair $(1, 8)$ ($1$ is a line, $8$ is formed of loops). Now, as per Equation~\ref{eq:policy}, filters in the layer are activated based on the channel-wise attention vector $v_L$ (see Equation~\ref{eq:fwd}), which are pushed together for examples of the same classes, and pushed away from each other for examples of different classes using prototype loss (Equation~\ref{eq:protoype_loss}). We visualize the t-SNEs of these $v_L$s on the test set in Figure~\ref{fig:clusters} and observe that the samples belonging to the same classes are clustered, confirming the effectiveness of our prototype loss. Moreover, the clusters of visually similar classes are close together, which is concomitant with the JSDs and class-wise activation rates seen earlier. Class similarities are also reflected through multiple clusters for the digit $9$, indicating its similarity with the digits $6$ (loop) and $1$ (line) in one cluster, but also with $7$ (line) and $4$ (line $+$ loop) in another cluster. Finally, we observe that there are examples that are scattered away from their class clusters and overlap with other clusters, probably indicating that these particular examples are visually closer to other digits. Note, however, that these similar examples and classes are distributed across tasks, which explains the lower similarities in activation patterns between task pairs in Figure~\ref{fig:taskwise} compared to the class pairs in Figure~\ref{fig:classwise}.

Therefore, \textit{Dynamos} is capable of learning modular and specialized units that result in input-adaptive dynamic separation and overlap of activations, based on the extent of similarities with previously learned examples. We also contend that the overlapping activations for digits of similar shape suggest the learning of general-purpose features.

\subsection{Trial-to-trial variability}
\label{subsubsec:variability}
\begin{figure}[t]
\centering
  \includegraphics[width=\linewidth]{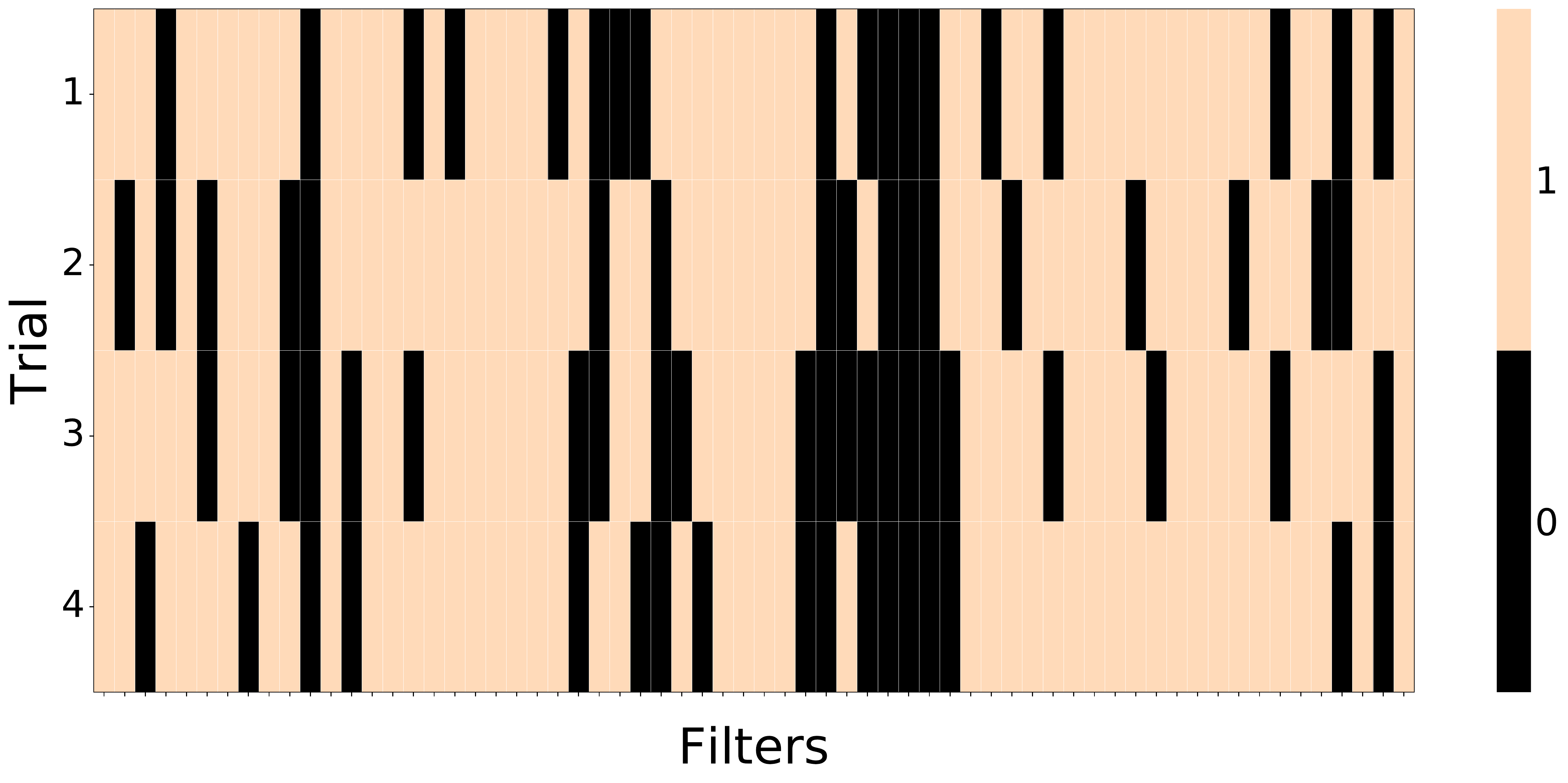}
  \caption{Trial-to-trial variability of responses to same input in \textit{Dynamos}. 
  } 
\label{fig:variability}
\end{figure}

The brain is known to show variability in response across trials~\cite{faisal2008noise,werner1963variability}. For the same stimulus, the precise neuronal response could differ between trials, a behavior absent in most conventional DNNs. In our method, this aspect of brains can be mimicked by using Bernoulli sampling instead of thresholding to pick keep/drop decisions at each convolutional layer.
In Figure~\ref{fig:variability}, we plot the response variability in the last convolutional layer of our DNN with the same example in four trials. We only pick responses for which the predictions were correct. It can be seen that each trial evoked a different response from the DNN. Furthermore, despite the differences, there are also some similarities in the response. 
There are some filters that are repeatedly left unused, as well as some filters that are used in every trial.
This demonstrates that \textit{Dynamos} can additionally simulate the trial-to-trial variability observed in brains.

\section{\uppercase{Conclusion and Future Work}}
\label{sec:conclusions}
We propose \textit{Dynamos}, a method for general continual learning, that simulates the dynamically modular and sparse response to stimuli observed in the brain. Dynamos rewards the input-adaptive removal of channel activations of convolutional layers using policy gradients for dynamic and sparse responses. 
To further induce modularity, channel-wise self-attention vectors corresponding to each convolutional layer are pulled together for examples from same classes, and are pushed apart for examples from different classes; these vectors are then used to sample the keep/drop decision for the corresponding channel. Using a memory buffer, we enforce multi-scale consistency between previous and current responses to prevent forgetting.
Dynamos outperforms previous baselines on multiple datasets when evaluated using class-incremental learning (CIL) and general continual learning (GCL) protocols. Dynamos exhibits similar and overlapping responses for similar inputs, yet distinct responses to dissimilar inputs by utilizing subsets of learned filters in an adaptive manner. We quantified the extent of class-wise overlaps and showed that the semantic similarity of classes (digits in MNIST, e.g. 1 and 7) are reflected in higher representation overlaps. We additionally visualized the channel-wise attention vectors and observed that they are clustered by the classes and the clusters of semantically similar classes lie together or overlap. Finally, we also demonstrated the ability of our method to mimic the trial-to-trial variability seen in the brain, where same inputs achieve same outputs through different ``responses", i.e. activations.
Thus, we consider our work as a step toward achieving dynamically modular and general-purpose continual learning. 


\bibliographystyle{apalike}
{\small
\bibliography{references}}

\section*{\uppercase{APPENDIX}}
\label{sec:appendix}

\begin{table}[h]
\centering
\caption{Class-Incremental learning accuracies for CCGN and Dynamos corresponding to Figure~\ref{fig:cil}.}
\label{tab:cil}
\begin{tabular}{ll|cc}
\hline
\begin{tabular}[c]{@{}l@{}}Buffer\\ Size\end{tabular} & Model & Seq-SVHN & Seq-MNIST \\ \hline
\multirow{2}{*}{500}  & CCGN  & 67.45    & 94.01 \\
                      & Dynamos  & 84.815   & 97.19 \\ \hline
\multirow{2}{*}{1000} & CCGN  & 73.99    & 95.94 \\
                      & Dynamos  & 87.38    & 97.51 \\ \hline
\multirow{2}{*}{2000} & CCGN  & 81.02    & 95.94 \\
                      & Dynamos  & 88.54    & 97.57 \\ \hline
\end{tabular}
\end{table}

\begin{table}[h]
\centering
\caption{Agent architecture with input features of shape $B\times C_{in}\times H\times W$, output features of shape $B\times C_{out}\times 1\times 1$, where $B$ is the batch size, $C_{in}$ is the number of channels in the input, and $C_{out}$ is the number of channels expected in the output, which is same as the number of keep/drop actions required for the corresponding convolutional layer.}
\label{tab:agent}
\resizebox{\columnwidth}{!}{%
\begin{tabular}{c|c|c}
\toprule
\textbf{Operations} & \textbf{Input size} & \textbf{Output size} \\
\midrule
Pointwise Conv  & $B\times C_{in}\times h\times w$   & $B\times C_{out}\times h\times w$ \\
Average Pooling & $B\times C_{out}\times h\times w$  & $B\times C_{out}\times 1\times 1$ \\
Reshape         & $B\times C_{out}\times 1\times 1$  & $B\times C_{out}$ \\
Linear          & $B\times C_{out}$                  & $B\times C_{out} / 16$ \\
ReLU            & $B\times C_{out} / 16$             & $B\times C_{out} / 16$ \\
Linear          & $B\times C_{out} / 16$             & $B\times C_{out}$ \\
Reshape         & $B\times C_{out}$                  & $B\times C_{out}\times 1\times 1$ \\
Sigmoid$_\tau$  & $B\times C_{out}\times 1\times 1$  & $B\times C_{out}\times 1\times 1$ \\ 
\bottomrule
\end{tabular}}
\end{table}

\begin{table*}[tb]
\centering
\caption{Architectures used in our experiments. For baseline experiments without dynamic compositionality, we do not use the "Agent" branch. Conv($k$, $n$, $s$, $p$) refers to convolutional layer with kernel size $k$, number of filters $n$, stride $s$, and padding $p$. BN refers to Batch Normalization. Linear($M$, $N$) refers to a linear layer with $M$-dimensional input and $N$-dimensional output. Agent($C_{in}$, $C_{out}$, $\tau$) refers to the Agent subnetwork with $C_{in}$ input channels, $C_{out}$ output channels, and $\tau$ temperature of the sigmoid in the probability layer (See Figure~\ref{fig:architecture}, Section~\ref{subsec:settings}). The elementwise multiplication of the actions from the agents with the output of a convolutional layer is done after the application of batch normalization, if present, but before the ReLU activation function, if present. For complete description of Agent architecture, refer to Table~\ref{tab:agent}. $num\_classes$ refers to the number of classes to be predicted.}
\label{tab:arch}
\begin{tabular}{c|c|c|c}
\toprule
\textbf{Component} & \textbf{Main Branch} & \textbf{Residual branch} & \textbf{Agent branch}\\
\midrule
Conv1 & $Conv(3, 32, 1, 1), BN, ReLU$ & $-$ & $-$ \\
\midrule
Block1 
    & $\left[\begin{array}{c}
     Conv(3, 32, 1, 1), BN, ReLU \\ 
     Conv(3, 32, 1, 1), BN, ReLU \end{array} \right]$ 
     & $\left[\begin{array}{c}
     Identity \\ Identity
     \end{array} \right]$ 
     & $\left[\begin{array}{c}
     Agent(64, 64, 0.15) \\ Agent(64, 64, 0.15)
     \end{array} \right]$ \\
\midrule
Block2 & 
     $\left[\begin{array}{c}
     Conv(3, 32, 1, 1), BN, ReLU \\ 
     Conv(3, 32, 1, 1), BN, ReLU 
     \end{array} \right]$ 
     & $\left[\begin{array}{c}
     Conv(1, 64, 2, 0), BN \\ 
     Conv(1, 64, 2, 0), BN 
     \end{array} 
     \right]$ 
     & $\left[\begin{array}{c}
     Agent(64, 64, 0.15) \\ 
     Agent(64, 64, 0.15) 
     \end{array} \right]$ \\
\midrule
Classifier & $Linear(64, num\_classes)$ & $-$ & $-$ \\
\bottomrule
\end{tabular}
\end{table*}

\begin{table*}[tb]
\begin{center}
\caption{Hyperparameters for all the datasets for Dynamos.}
\label{tab:settings}
\resizebox{\textwidth}{!}{%
\begin{tabular}{lc|cccccccccccc}
\toprule
Dataset & \begin{tabular}[c]{@{}c@{}}Buffer\\ Size\end{tabular} & lr & \#Epochs & \begin{tabular}[c]{@{}c@{}}\#Warmup\\ Ep./Itr.\end{tabular} &\begin{tabular}[c]{@{}c@{}}Batch\\ Size\end{tabular} & \begin{tabular}[c]{@{}c@{}}Memory\\Batch\\ Size\end{tabular} & $\alpha$ & $\beta$ & $\alpha_P$ & $\lambda$ & $\gamma$ & $w_P$ & $kr$ \\ \midrule
\multirow{3}{*}{{Seq-MNIST}} & $500$ & $0.07$ & $1$ & $10$ it & $10$ & $10$ & $0.2$ & $2.0$ & $0.2$ & $500$ & $0.5$ & $0.3$ & $0.7$\\
 & $1000$     & $0.07$ & $1$ & $10$ it & $10$ & $10$ & $0.1$ & $2.5$ & $0.2$ & $200$ & $0.7$ & $0.5$ & $0.7$\\ 
 & $2000$     & $0.07$ & $1$ & $10$ it & $10$ & $10$ & $0.5$ & $3.0$ & $0.2$ & $200$ & $0.5$ & $0.5$ & $0.7$\\ \midrule
\multirow{3}{*}{{SVHN}} & $500$ & $0.07$ & $70$ & $10$ ep & $16$ & $16$ & $2.0$ & $3.0$ & $1.0$ & $500$ & $1.0$ & $0.5$ & $0.7$ \\
 & $1000$     & $0.07$ & $70$ & $10$ ep & $16$ & $16$ & $2.5$ & $2.0$ & $0.2$ & $500$ & $1.0$ & $0.5$ & $0.7$ \\ 
 & $2000$     & $0.07$ & $70$ & $10$ ep & $16$ & $16$ & $2.5$ & $2.0$ & $0.2$ & $500$ & $1.0$ & $0.5$ & $0.7$\\ \midrule
\multirow{3}{*}{{MNIST-360}} & $100$ & $0.07$ & $1$ & $10$ it & $16$ & $16$ & $0.2$ & $1.0$ & $0.1$ & $200$ & $0.5$ & $0.5$ & $0.7$ \\
 & $200$     & $0.07$ & $1$ & $10$ it & $16$ & $16$ & $0.2$ & $1.5$ & $0.1$ & $200$ & $1.0$ & $0.5$ & $0.7$\\
 & $500$     & $0.07$ & $1$ & $10$ it & $16$ & $16$ & $0.1$ & $1.5$ & $0.1$ & $200$ & $0.3$ & $0.3$ & $0.7$ \\
\bottomrule
\end{tabular} }
\end{center}
\end{table*}

\begin{figure*}[tb]
\begin{tabular}{cc}
\subfloat[][Classwise activations for Layer 1]{\includegraphics[width=0.48\linewidth]{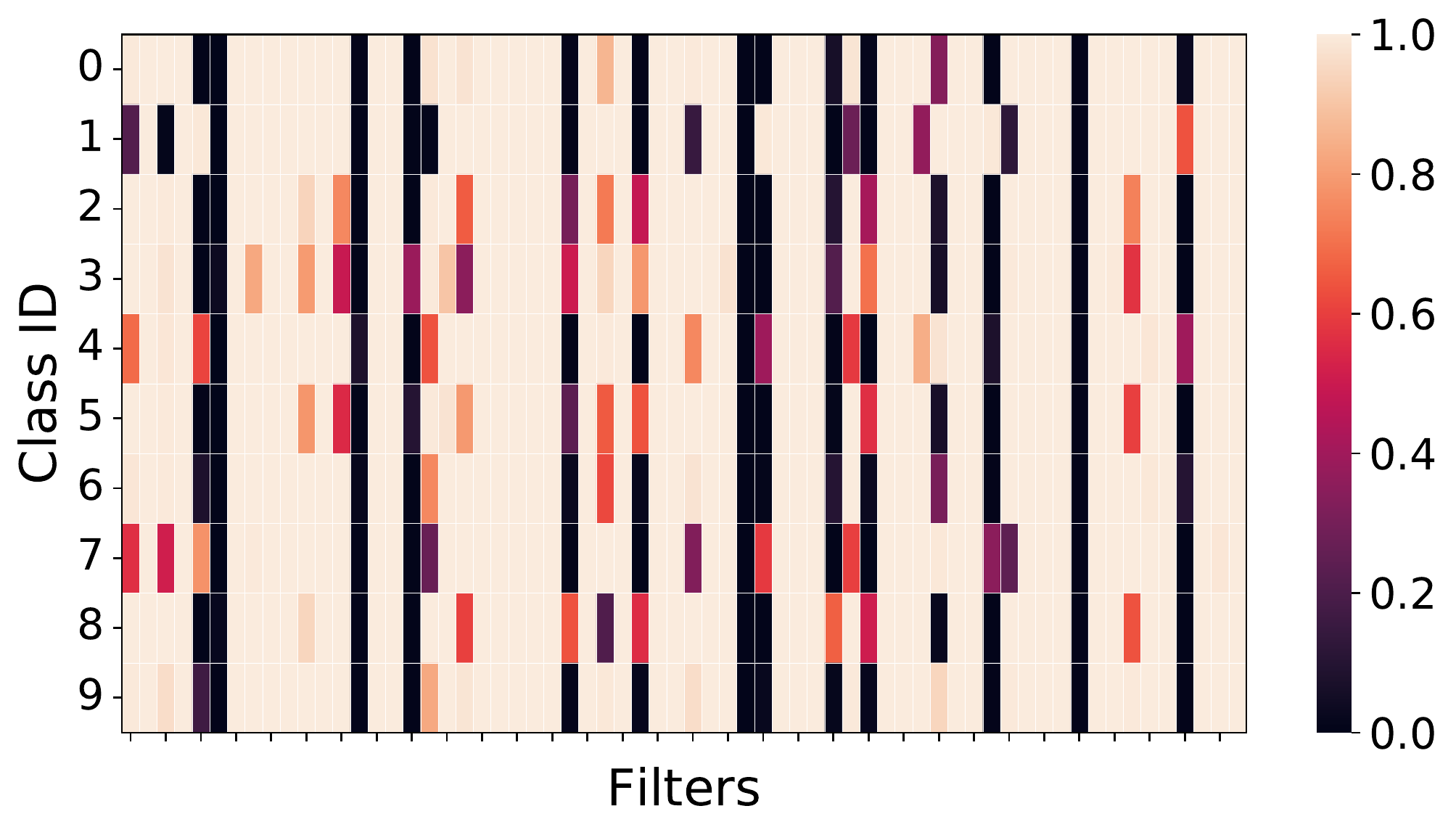}} &
\subfloat[][Taskwise activations for Layer 1]{\includegraphics[width=0.48\linewidth]{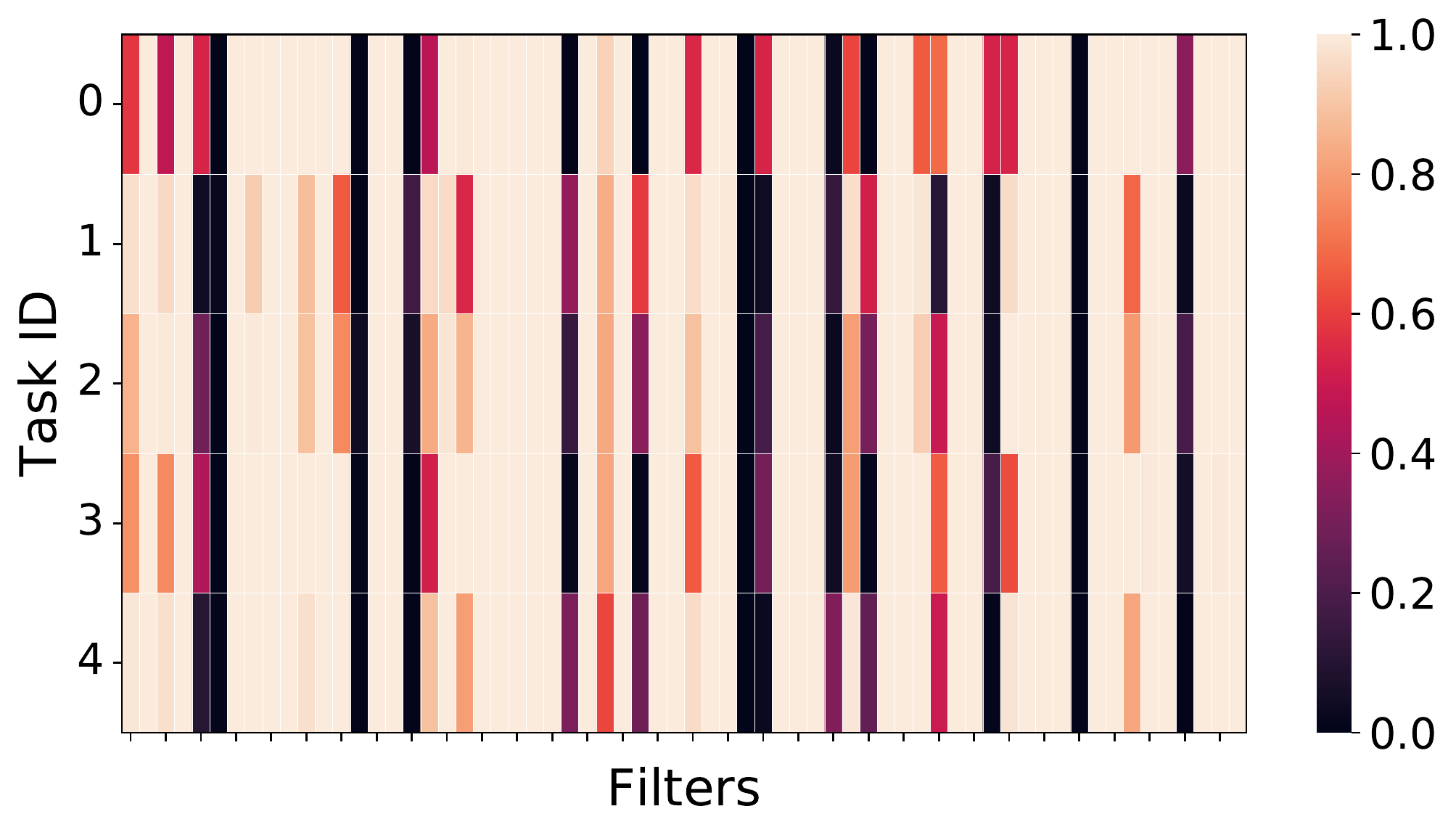}} \\ \\
\subfloat[][Classwise activations for Layer 2]{\includegraphics[width=0.48\linewidth]{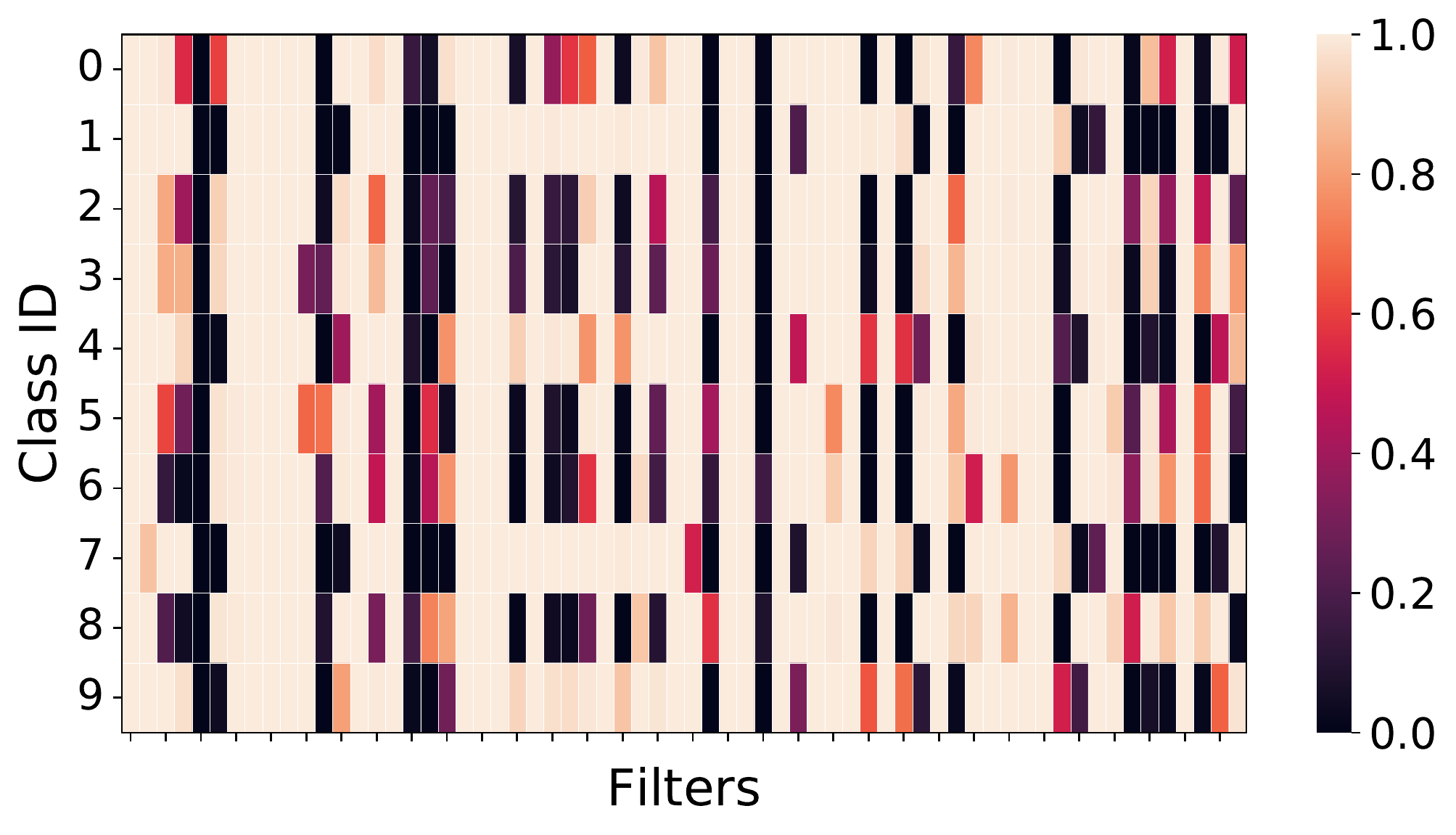}} &
\subfloat[][Taskwise activations for Layer 2]{\includegraphics[width=0.48\linewidth]{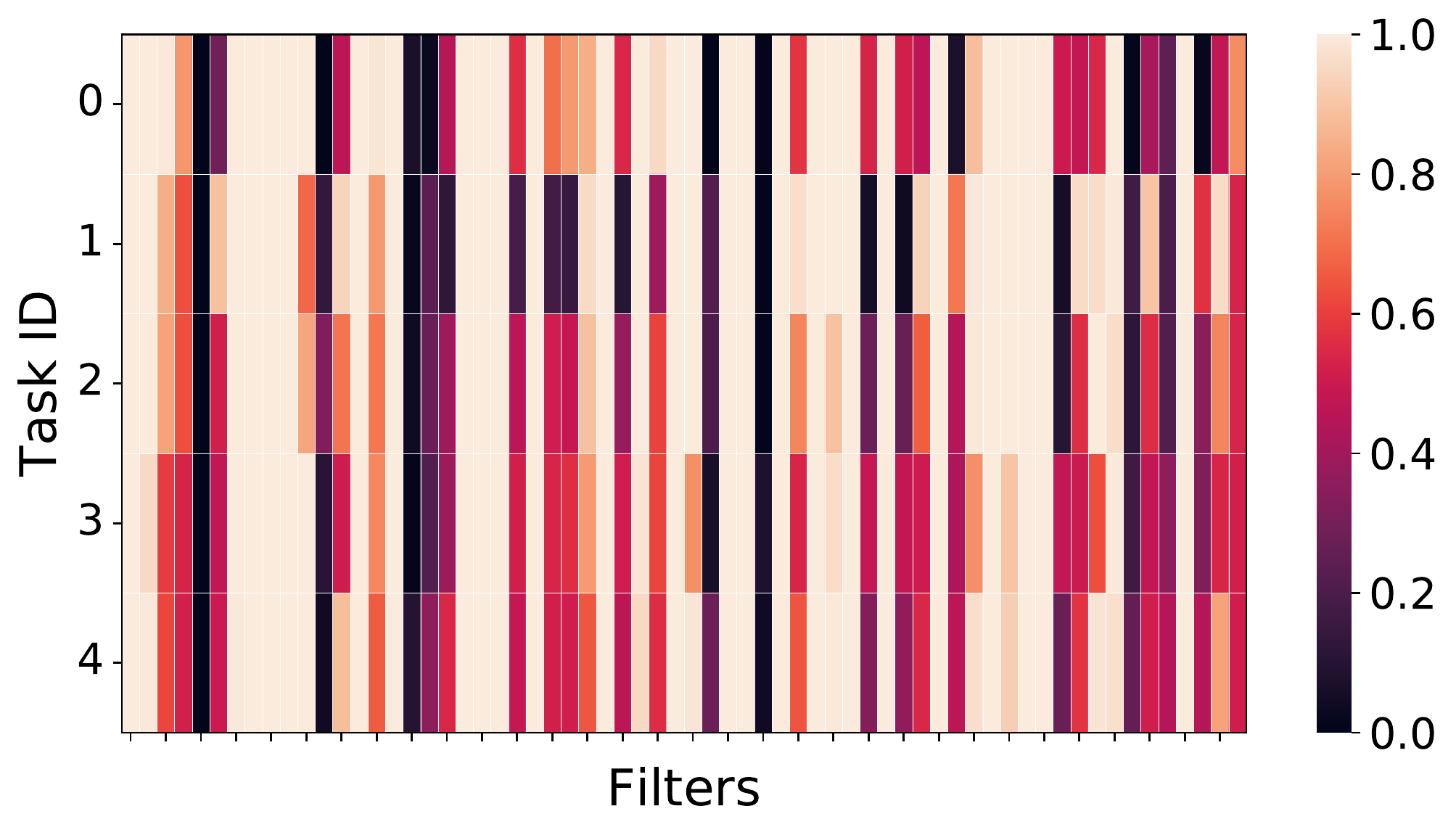}} \\ \\
\subfloat[][Classwise activations for Layer 3]{\includegraphics[width=0.48\linewidth]{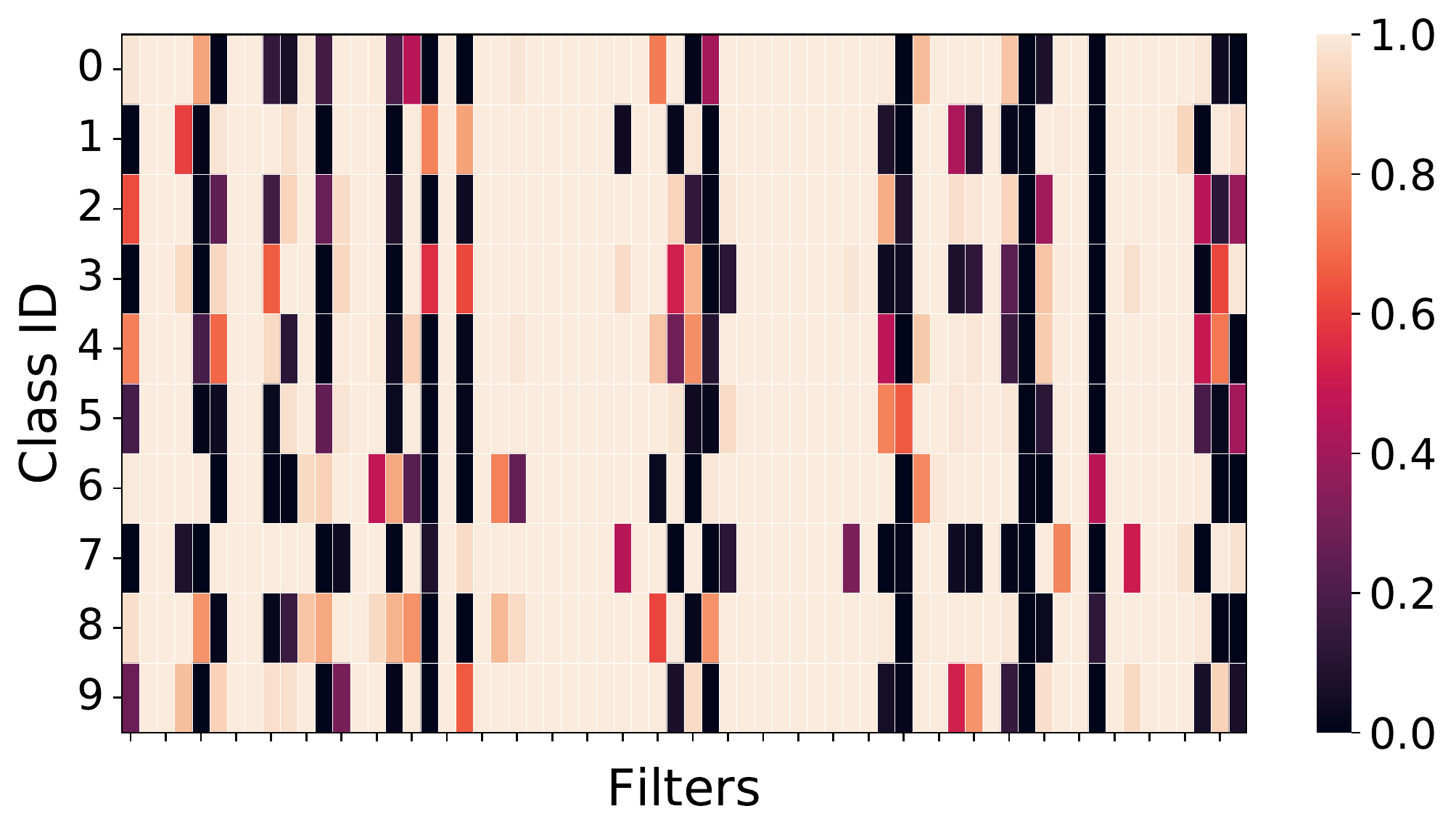}} &
\subfloat[][Taskwise activations for Layer 3]{\includegraphics[width=0.48\linewidth]{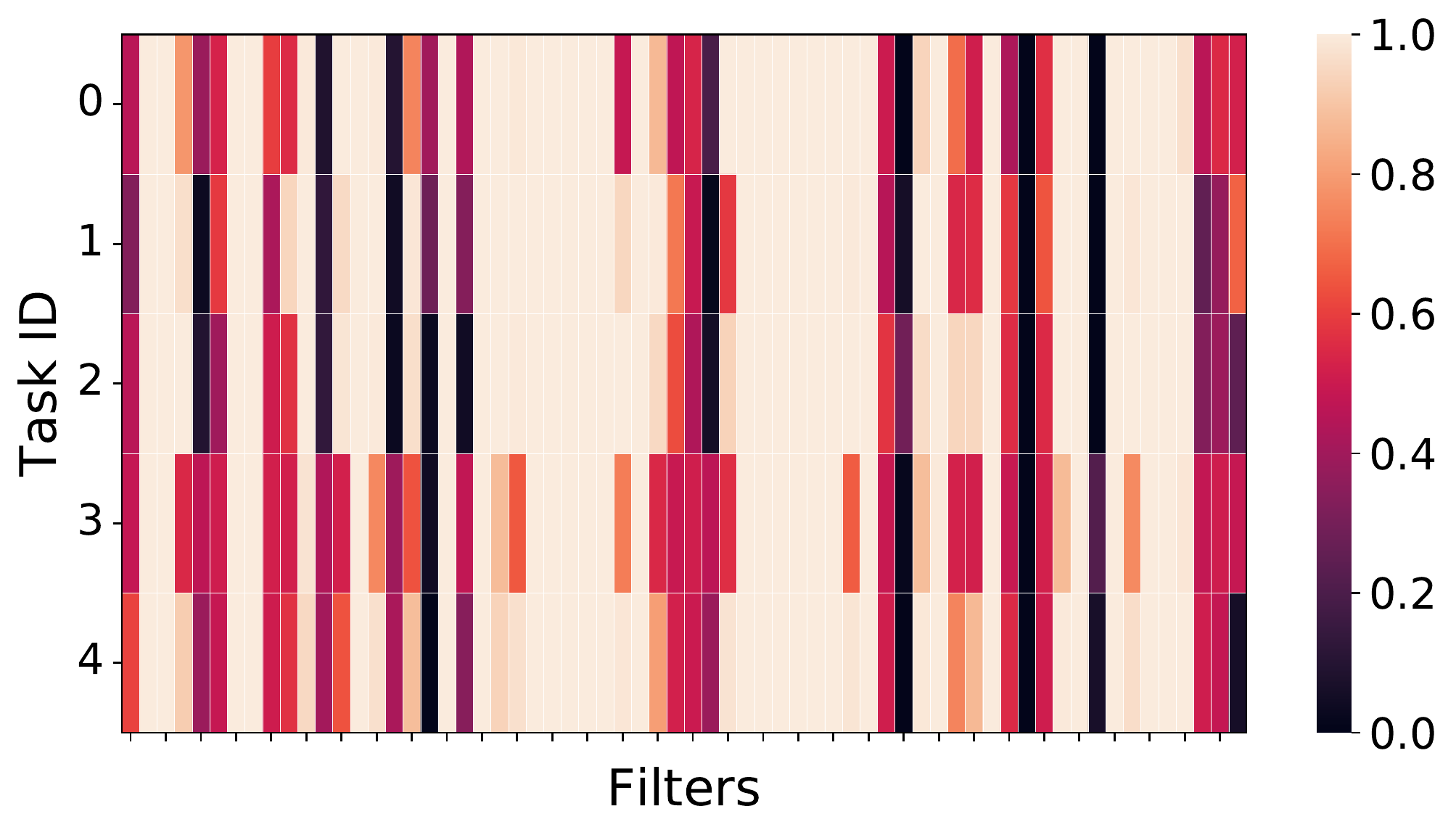}} \\ \\
\subfloat[][Classwise activations for Layer 4]{\includegraphics[width=0.48\linewidth]{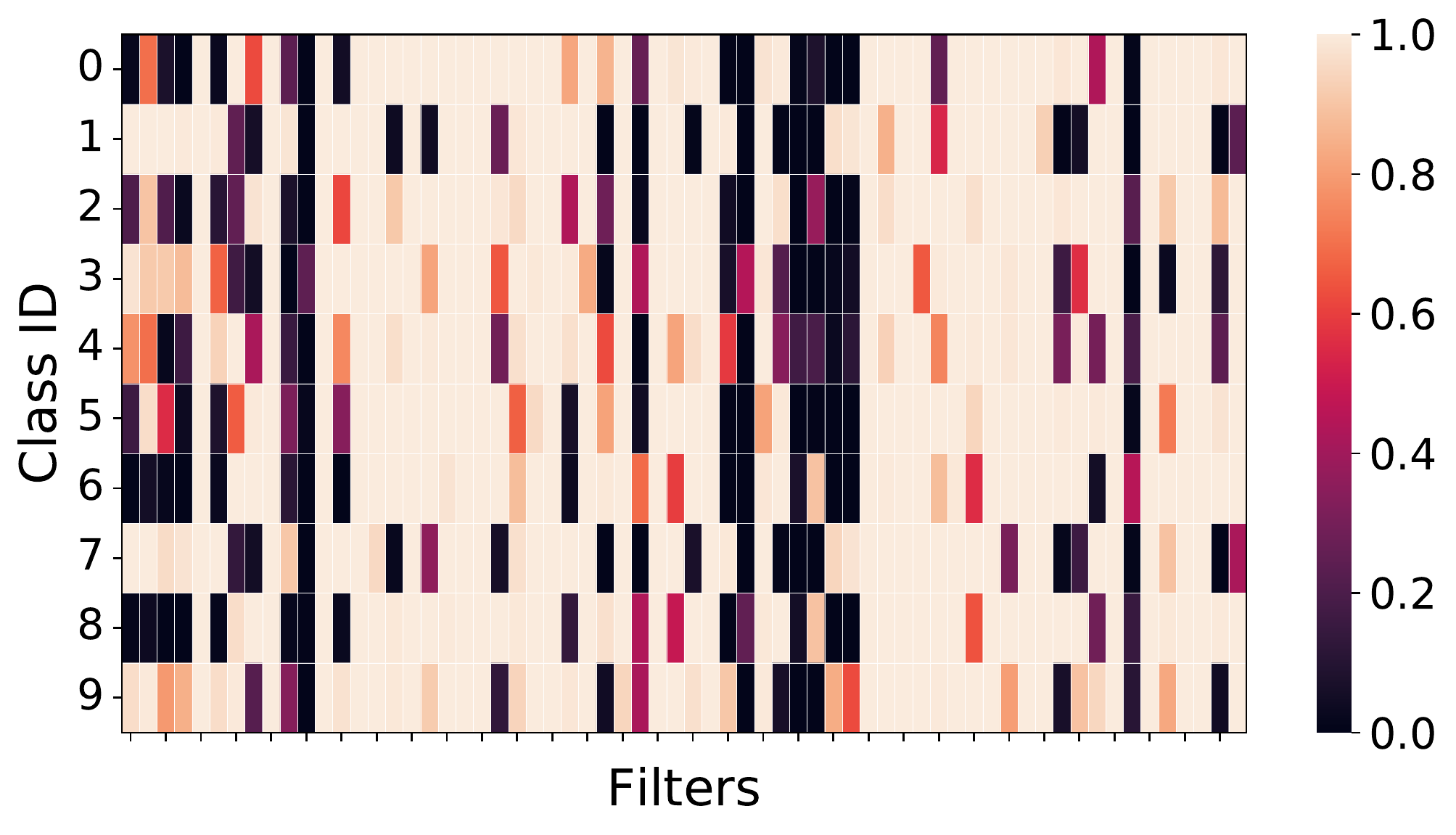}} &
\subfloat[][Taskwise activations for Layer 4]{\includegraphics[width=0.48\linewidth]{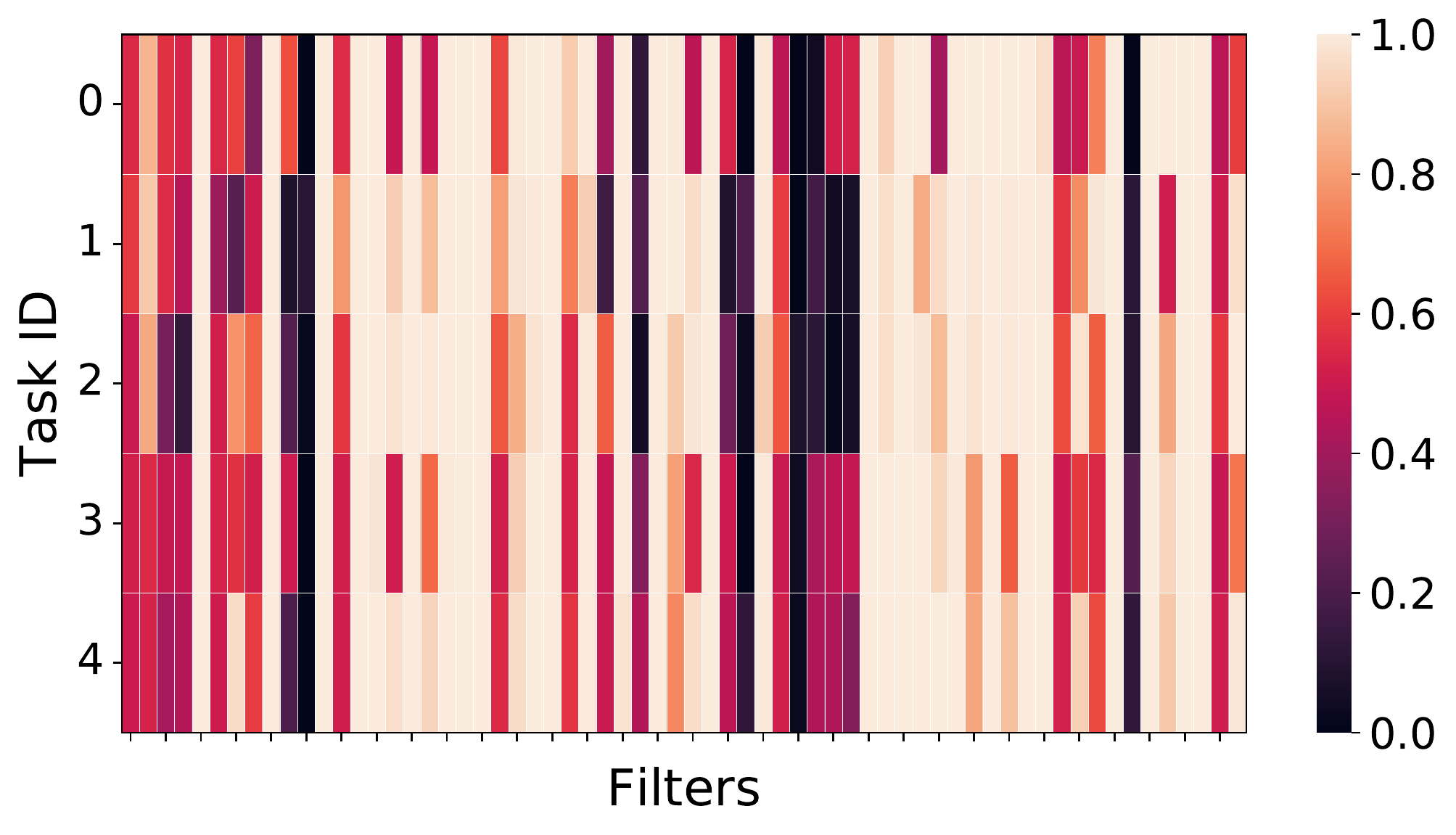}} \\ \\
\end{tabular}
\caption{Filter activation rates for each filter in each convolutional layer of Block 2 with respect to MNIST tasks and classes. Overlapping activations of tasks and classes indicative of similarities between them can still be observed. For e.g. 1 and 7 still show very similar responses.} 
\label{fig:modularity_full}
\end{figure*}

\end{document}